\documentclass{article}

    \PassOptionsToPackage{numbers, compress}{natbib}

     \usepackage[preprint]{neurips_2023}

\usepackage[utf8]{inputenc} 
\usepackage[T1]{fontenc}    
\usepackage{hyperref}       
\usepackage{url}            
\usepackage{booktabs}       
\usepackage{amsfonts}       
\usepackage{nicefrac}       
\usepackage{microtype}      
\usepackage{xcolor}         
\usepackage{multicol}
\usepackage{amsmath}
\usepackage{amssymb}
\usepackage{enumitem}
\usepackage{colortbl}
\usepackage{subcaption}
\usepackage{pifont}
\usepackage{fancyvrb}
\usepackage{siunitx}
\usepackage{duckuments}
\usepackage{graphicx}
\usepackage{wrapfig}
\usepackage{transparent}

\definecolor{maroon}{cmyk}{0,0.87,0.68,0.32}
\definecolor{customgreen}{cmyk}{1,0,1,0.5}

\newcommand{\Ours}{ZeroAvatar}
\newcommand{\bm}[1]{\mathbf{#1}}

\usepackage[font=small,labelfont=bf]{caption}  
\title{ZeroAvatar: Zero-shot 3D Avatar Generation from a Single Image}

\author{%
  Zhenzhen Weng, Zeyu Wang, Serena Yeung\\
  Stanford University\\
  {\tt\small \{zzweng, wangzeyu, syyeung\}@stanford.edu }
}

\begin{document}

\maketitle
\begin{abstract}
Recent advancements in text-to-image generation have enabled significant progress in zero-shot 3D shape generation. This is achieved by score distillation, a methodology that uses pre-trained text-to-image diffusion models to optimize the parameters of a 3D neural presentation, e.g. Neural Radiance Field (NeRF). While showing promising results, existing methods are often not able to preserve the geometry of complex shapes, such as human bodies.  To address this challenge, we present ZeroAvatar, a method that introduces the explicit 3D human body prior to the optimization process. Specifically, we first estimate and refine the parameters of a parametric human body from a single image. Then during optimization, we use the posed parametric body as additional geometry constraint to regularize the diffusion model as well as the underlying density field. Lastly, we propose a UV-guided texture regularization term to further guide the completion of texture on invisible body parts. We show that ZeroAvatar significantly enhances the robustness and 3D consistency of optimization-based image-to-3D avatar generation, outperforming existing zero-shot image-to-3D methods. 
\end{abstract}
\section{Introduction}
The ability to extract rich and accurate 3D information from a single image holds great importance in content creation, where realistic and immersive visual experiences are crucial. By automatically inferring the 3D structure and appearance of objects, artists and designers can efficiently generate lifelike virtual scenes, characters, and objects. Beyond the realms of content creation and AR/VR, the ability to perceive 3D geometry and appearance from a single image has broader implications, and plays a pivotal role in robotics and scene understanding \cite{weng2021holistic,yin2021learning}. Despite the recent advancements in computer vision, 3D perception from a single image remains a challenging task. This is primarily due to the loss of information in the image projection process, which obscures important depth cues and object characteristics. Consequently, researchers have proposed various approaches and algorithms to tackle this problem, aiming to extract accurate and detailed 3D information from a single image.

\begin{figure}[t]
    \centering
    \includegraphics[width=\textwidth]{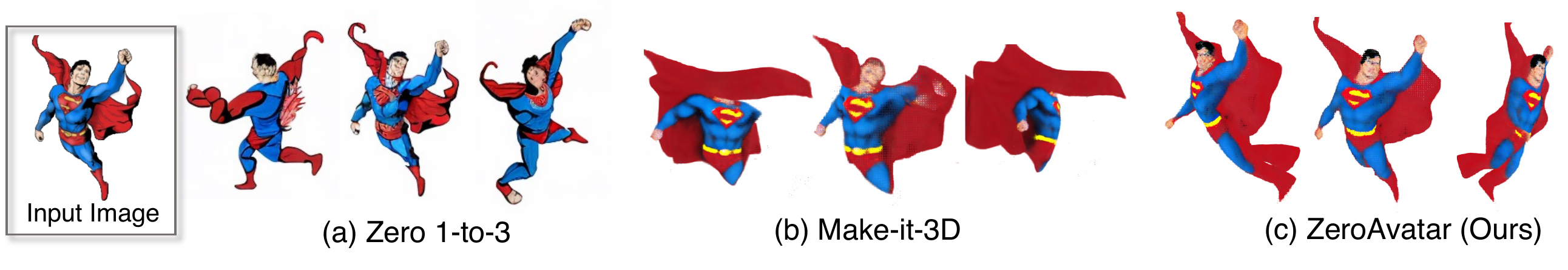}
    \caption{We propose \Ours{}, a zero-shot method that generates high-fidelity 3D avatars from single-view images. \Ours{} significantly improves over existing zero-shot methods in preserving the human structure. Zero 1-to-3 predicts 2D views from novel angles, whereas Make-it-3D and \Ours{} output a 3D model.}
    \label{fig:teaser}
\end{figure}

3D reconstruction from a single image has traditionally been approached through learning-based methods \citep{wang2018pixel2mesh,gkioxari2019mesh,xiu2022econ}.
These approaches involve training neural networks to map an input image to a corresponding 3D representation. However, a challenge with this approach is the scarcity of high quality 3D training data. Recently, driven by the advancements in Large Language Models (LLMs) \cite{radford2021learning} and text-to-2D generative models \citep{saharia2022photorealistic,nichol2021glide,Rombach_2022_CVPR} that are trained on large-scale data, a new avenue of exploration has opened up, offering the possibility of zero-shot generation for 3D representations. These type of approaches \citep{liu2023zero,seo2023let,tang2023make,xu2022dream3d,poole2022dreamfusion} harnesses the prior information embedded in pre-trained models to optimize the parameters of implicit 3D representations of the object, leading to enhanced fidelity of the reconstructed 3D geometry and appearance. Despite showing impressive results on a vast variety of objects, these approaches have a notable limitation when it comes to accurately reconstructing humans with complex poses (Figure \ref{fig:teaser}). This is because these methods do not explicitly model the structure of human bodies.


In this work, we present \Ours{}, a zero-shot 3D generation method that generates a high-fidelity 3D avatar from a single image. We address the limitations of existing works by employing several strategies. First, we initiate the average density field of neural radiance field (NeRF) \cite{mildenhall2021nerf} by deriving a rough shape from an estimated human body shape (in the form of a parametric body model). Subsequently, we employ the depth information obtained from the posed body model as an extra conditioning for the text-to-image model (i.e. Stable Diffusion \citep{Rombach_2022_CVPR}), which enables us to generate results that better align with the geometric characteristics of the posed human. Furthermore, we incorporate a UV-guided texture prior to regularize the appearance of invisible points. We show that \Ours{} effectively enhances the overall fidelity and realism of the generated humans.

Our main contributions can be summarized as follows:
\begin{itemize}
    \item We propose \Ours{}, a method for creating high-fidelity 3D avatars from a single image, using a pre-trained text-to-image diffusion model as a prior. 
    \item By incorporating SMPL body model as an explicit geometry prior, along with a depth-conditioned score distillation loss and a UV-guided prior for invisible body parts, \Ours{} significantly improves both geometry and appearance of the generated avatars, surpassing existing state-of-the-art zero-shot 3D generation techniques.
    \item \Ours{} enables applications such as zero-shot text-to-3D avatar generation. We show that using the generated image from a pre-trained text-to-image model as a stepping stone, \Ours{} is able to generate 3D avatars with pose or text control, allowing for a wide range of downstream applications.
\end{itemize}
\section{Related Work}
\subsection{3D Generation from 2D Observations}
Inferring 3D geometry from multi-view data is a classical computer vision problem. Earlier approaches perform multi-view or multi-modal fusion, which involves combining information from multiple views or modalities to generate more accurate 3D representations. However, completeness of the observations is essential in this type of approach, which would require dense observations of the scene from various angles. In recent years, there have been significant breakthroughs in this field enabled by implicit neural representations (e.g. neural radiance fields (NeRF))\citep{sitzmann2019scene,mildenhall2021nerf}, which reduces the need for dense observations. For articulated objects (e.g. humans) in particular, prior works learn pose-conditioned NeRF from 2D observations, e.g. a monocular video \citep{su2021nerf,weng2022humannerf}, or sparse 2D observations \citep{zhao2021humannerf}. 

\subsection{3D Humans from a Single Image}
The task of estimating human pose and shape from a single image is referred to as single-view Human Mesh Recovery (HMR). There are typically two approaches for model-based HMR: learning-based and optimization-based. Learning-based methods \citep{kanazawa2018end} train an HMR model end-to-end on large human pose datasets, and can be bottle-necked by the scarcity of training data paired with 3D ground truths. Consequently, for challenging out-of-domain scenarios, domain adaptation \citep{guan2021bilevel,weng2022domain} is often needed to close the domain gap. On the other hand, optimization-based methods \citep{bogo2016keep,SMPL-X:2019} employ an iterative fitting routine to estimate the body pose and shape of a parametric human body model that best explains 2D observations, e.g. 2D joint locations or human silhouettes. Since explicitly optimizing for the agreement of the model with image features, the body model ends up having good alignment with the reference image. 

Besides the model-based human recovery, there is also a line of work that directly learns the volumetric representation of a human from single-view images \citep{saito2019pifu,xiu2022icon,xiu2022econ}. These methods require end-to-end training, and therefore have limited generalization capability due to the scarcity of available 3D human scans. \Ours{} is analogous to the optimization-based model-based HMR methods in that we optimize a 3D representation of the human body to achieve consistency with the input image. In contrast, our method extends beyond model-based methods by recovering surface details such as clothing and closely-interacting objects (e.g. basketball in hand). Compared to learning-based methods that predict volumetric representation from a single image, our method exhibits a broader range of generalization, demonstrating superior performance on real-world humans as well as virtual avatars, such as cartoon characters. The ability to handle such a broad range of humans expands the potential applications and creative possibilities of our method in the realm of virtual character generation and animation.

\subsection{Zero-Shot 3D Generation}
Existing zero-shot 3D generation methods typically employ pre-trained vision-language models. Earlier works such as Text2Mesh \citep{michel2022text2mesh} and DreamFields \citep{jain2022zero} use pre-trained CLIP model \citep{radford2021learning} as guidance to optimize the appearance and geometry of the 3D representation, e.g. mesh or neural radiance field (NeRF). Analogously for human subjects, CLIP-actor \citep{youwang2022clip} and AvatarCLIP \citep{hong2022avatarclip} employ CLIP-guided losses for generating stylized human avatars based on text descriptions.

A more recent type of approach (e.g. DreamFusion \citep{poole2022dreamfusion}, Magic3D \citep{lin2022magic3d}, Score Jacobian Chaining \citep{wang2022score}) have made significant advancements in pushing the quality of generated 3D representations to new levels leveraging large diffusion models \citep{saharia2022photorealistic,Rombach_2022_CVPR}. These works leverage the idea of Score Distillation Sampling (SDS), where a score distillation loss derived from pre-trained text-to-image diffusion models (e.g. \citep{saharia2022photorealistic}) is used as a prior for optimization, such that the 2D renderings from random angles look like good generations from text-to-image models. Following these pioneering works, recent works \cite{tang2023make,liu2023zero} have approached the problem of zero-shot image-conditioned 3D generation. Although these methods are able to recover high-fidelity 3D representation of objects with simple geometry, they often struggle to hallucinate the geometry of relatively more complex structure. To overcome this limitation, an increasing number of works \cite{liu2022iss,xu2022dream3d,seo2023let} are starting to utilize shape priors as a more robust guide to aid the learning of geometry. In creating shape priors, ISS \citep{liu2022iss} and Dream3D \citep{xu2022dream3d} utilize a fine-tuned image-to-mesh model, and 3D Fuse \citep{seo2023let} leverages an off-the-shelf model to generate a coarse point cloud of the scene. 
\begin{figure}[t!]
    \centering
    \includegraphics[width=\textwidth]{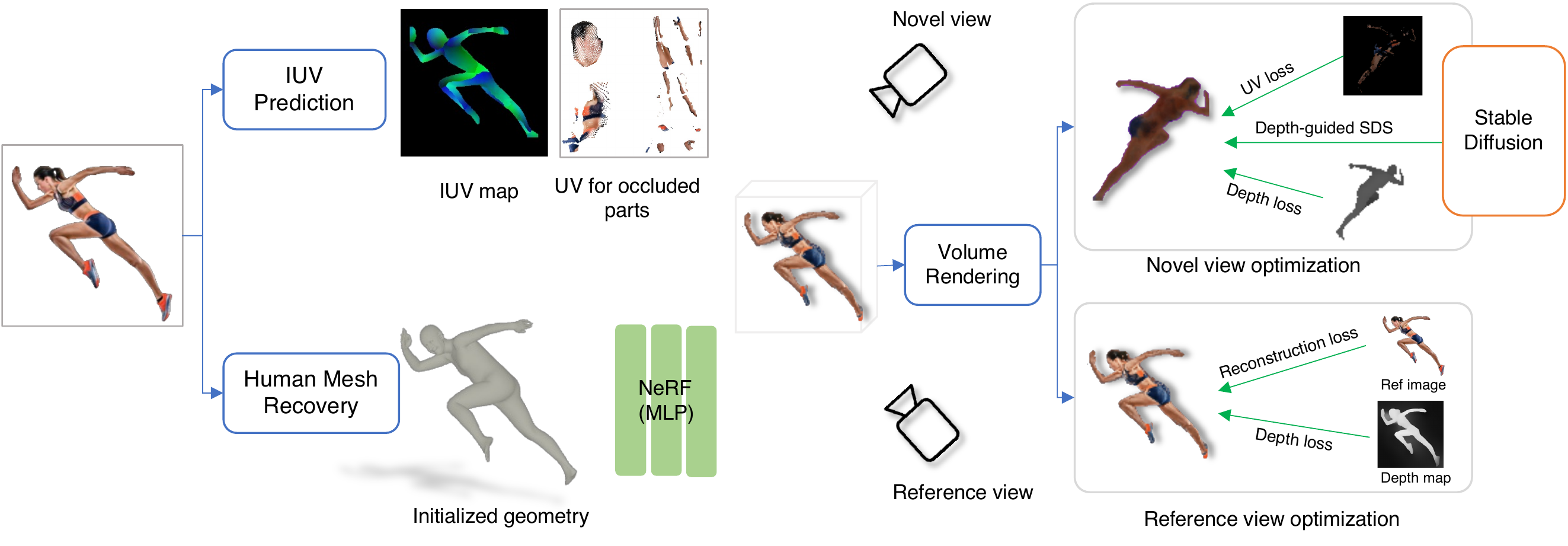}
    \caption{Overview of \Ours{}. Given a single image, we first estimate the body pose, shape, and UV map of the person. We use the estimated body mesh to initialize the density field of the 3D representation. Then, we optimize the appearance and refine the geometry of the person using Score Distillation Sampling, during which depth information from the posed body model is used as conditioning in addition to text (i.e. image caption). When optimizing given a sampled novel view, we additionally use the inferred UVs on the invisible body parts to aid the learning of appearance.}
    \label{fig:sys_figure}
\end{figure}

\section{\Ours{}}
Given a reference image $\mathcal{I}$ of a human avatar, \Ours{} optimizes a neural radiance field (NeRF) \citep{mildenhall2021nerf} that represents the 3D avatar. To achieve this, we first initialize the mean density field of NeRF with a coarse shape computed from a parametric body model (Section \ref{sec:density_init}). Then, we use the depth of the posed body model as additional conditioning for the Stable Diffusion model to produce generations that are more faithful to the geometry of the posed human (Section \ref{sec:depth_guided_opt}). In addition, we regularize the appearance of invisible points using a UV-guided texture prior leveraging a common property of human textures (Section \ref{sec:uv_opt}).

\subsection{Background}
\label{sec:background}
\textbf{Neural radiance field (NeRF) \citep{mildenhall2021nerf}} represents a 3D scene via an implicit function
\begin{align}
    F_{\theta}(\gamma(\bm{x})) = (\sigma(\bm{x}), c(\bm{x}))
\end{align}
where $\gamma(\cdot)$ is a frequency encoder, and $\sigma$ and $c$ are density and colors and can be learned by a small MLP with parameters $\theta$. 
We render a neural field using the volume rendering equation from \citet{mildenhall2021nerf}. For each image pixel, a ray $\bm{r}$ is casted from the pixel location into the 3D scene and the RGB value at the pixel location can be calculated using the density and color values (predicted by $F_{\theta}$) from the $D$ sampled 3D points $x_i$ along $\bm{r}$. Formally, color $C(\bm{r})$ can be expressed as
\begin{align}
    C(\bm{r}) = \sum_{i=1}^{D} W_i (\alpha_i \Pi_{j<i} (1-\alpha_j)) c(\bm{x_i})
\end{align}
where $\alpha_i = 1- exp({-\sigma (\bm{x_i}) \Delta_i})$, and $\Delta_i = ||\bm{x_i} -\bm{x_{i+1}}||$ is the interval between sample i and i+1.

\textbf{Skinned Multi-Person Linear (SMPL) \citep{loper2015smpl}} body model is a differentiable function $\mathcal{M}(\Omega, \beta, t)$ that takes a pose parameter (i.e. $K$ joint rotations along the kinematic tree) $\Omega \in \mathbb{R}^{K \times 3}$, shape parameter $\beta \in \mathbb{R}^{10}$ and a 3D translation vector $\mathbb{R}^{3}$, and returns the body mesh $\mathcal{M} \in \mathbb{R}^{6890 \times 3}$ with $6890$ vertices. We use SMPL as a 3D human body shape prior in \Ours{}.

\subsection{Density Field Initialization}
\label{sec:density_init}
The goal of this stage is to initialize the density field with a reasonable human shape. This provides a reasonable starting point for the geometry of the person, and prevents the optimization from diverging.

Given an input image, we first estimate the pose and shape of the person using an off-the-shelf human mesh recovery (HMR) model \cite{zhang2021pymaf}. The proposed optimization losses in Section \ref{sec:depth_guided_opt} and \ref{sec:uv_opt} assume good alignment between the SMPL body and the image. Direct SMPL mesh estimation $\mathcal{M}$ may not align well with the reference image. Therefore, using a similar loss term as in \citet{xiu2022icon}, we refine the predicted SMPL parameters by encouraging the consistency between the SMPL prediction, normal map and silhouette. The loss function for the refinement is
\begin{align}
    \mathcal{L}_{\text{SMPL}} &= min_{\Omega, \beta, t} (\lambda_{\mathcal{N}} |\mathcal{N} - \mathcal{N}^{\mathcal{M}}| + \lambda_{\mathcal{S}}|\mathcal{S} - \mathcal{S}^{\mathcal{M}}|) \label{eq:refine}
\end{align}
where $\mathcal{N}^{\mathcal{M}}$ and $\mathcal{S}^{\mathcal{M}}$ are the normals and silhoutte of the predicted mesh $\mathcal{M}$, $\mathcal{N}$ and $\mathcal{S}$ are the normals and mask of the human in the image. $\lambda_{\mathcal{N}}$ and $\lambda_{\mathcal{S}}$ are hyper-parameters.

Next, we use the refined human mesh $\mathcal{M}^{*}$ to initialize the neural radiance field. Specifically, for each point $\bm{x}$ in space, we compute the signed distance between $\bm{x}$ and the closest point on SMPL surface. Following \citet{xu2022dream3d}, we then initialize the density field using the signed distance $d(x, \mathcal{M}^{*})$.
\begin{align}
    v_{x} &= \frac{1}{\beta} \text{sigmoid}(- \frac{d(x, \mathcal{M}^{*})}{\beta}) \\
    \bar{\sigma_{x}} &= \text{max}(0, \text{softplus}^{-1} (v_{x}))
\end{align}
$\beta$ is a hyper-parameter controlling the sharpness of the shape boundary, and we set it to $0.05$. In optimizing NeRF, we use a MLP to predict the residual density at a point $\bm{x}$. That is,
\begin{align}
    (\sigma(\bm{x}), c(\bm{x})) = F_{\theta}(\gamma(\bm{x})) + (\bar{\sigma_{x}}, \bm{0})
\end{align}

\subsection{Depth-guided Geometry Optimization 
} 
\label{sec:depth_guided_opt}
In this stage, we optimize the neural radiance field leveraging a pre-trained text-to-image diffusion model (i.e. Stable Diffusion \cite{Rombach_2022_CVPR}). Diffusion models are generative models that are trained to invert a multi-step noising process. The denoising process at each step is learned by a neural network $\epsilon_{\phi}$. To optimize the 3D representation of the human such that the renderings are close to the good generation samples, \citet{poole2022dreamfusion} proposed Score Distillation Sampling (SDS). 

Given a novel view rendering $\bm{x} = \mathcal{G}_{\theta}(\beta)$ of the 3D representation from viewpoint $\beta$ ($\mathcal{G}$ is the volume rendering function as described in Section \ref{sec:background}), SDS optimizes the parameters of NeRF $\theta$ via the objective
    $\nabla_{\theta}\mathcal{L}_{SDS}(\phi, \bm{x} = \mathcal{G}_{\theta}(\beta)) = \mathbb{E}_{t, \bm{\epsilon}} [w(t) (\bm{\epsilon}_{\phi}(\bm{z}_t; y, t) - \bm{\epsilon}) \frac{\partial \bm{z}}{\partial \bm{x}} \frac{\partial\bm{x}}{\partial \theta}] $.
Here $y$ is the text embedding of the reference image caption, and $\bm{z}_t$ is the noisy latent of the novel view, obtained by the image encoder of Stable Diffusion. While SDS is effective in optimizing scenes that have a simple geometry such as a blob, we observe that for objects with complex geometry such as a human with stretched out limbs, the vanilla SDS alone often fails to preserve the underlying structure of the scene. In this work, we leverage depths of the SMPL mesh as a geometric prior for Stable Diffusion to regularize the structure of the generated images. Specifically, we first use a depth renderer $\mathcal{R}_d$ to render the depth values of SMPL mesh from viewpoint $\beta$ to get a depth map $d = \mathcal{R}_d(\mathcal{M}, \beta)$, and then use the depth-conditioned denoising model $\bm{\epsilon}_{\phi'}^{d}$ (a variant of $\bm{\epsilon}$ that takes an additional depth channel). The depth-conditioned score distillation objective is then
\begin{align}
    \nabla_{\theta}\mathcal{L}_{d-SDS}(\phi, \bm{x} = \mathcal{G}_{\theta}(\beta)) = \mathbb{E}_{t, \bm{\epsilon}} [w(t) (\bm{\epsilon}^{d}_{\phi'}(\bm{z}_t; y, t, d) - \bm{\epsilon}) \frac{\partial \bm{z}}{\partial \bm{x}} \frac{\partial\bm{x}}{\partial \theta}]
\end{align}
With the density initialization and depth-conditioned SDS, geometry of the person improves quickly in beginning epochs of the optimization. However, since SMPL body model does not include surface details such as clothing and hair, conditioning on SMPL depth throughout training would hurt the optimization if the person's surface is far from the body surface. Hence, we utilize SMPL depth conditioning for the first 10 epochs of the optimization, and then switch to using the predicted depth map as conditioning so that clothing details are preserved. 

\subsection{UV-Guided Texture Completion}
\label{sec:uv_opt}
Since we are using a parametric mesh model as a proxy of the 3D representation, it is advantageous to leverage one of its inherent benefits, which is the existing correspondence between vertices across the views enabled by UV maps. Notably, in the case of human meshes, the textures often exhibit symmetrical patterns between their left and right counterparts. Motivated by these two observations, we introduce a term to facilitate the texture learning of the occluded body part. The term is especially useful when the side of the person is visible, in which case score distillation loss alone often results in blurry textures on the occluded region. With the predicted back textures generated by the reflection of the visible part, the back textures get better prior, and the final results have better appearance on the invisible part (Figure \ref{fig:ablations}).

We first use DensePose \citep{guler2018densepose} to regress the UV coordinates from the reference image, and sample the RGB colors from the reference image to fill in the visible region of the UV map $\mathcal{T}_{\text{visible}}$. Then, we find the symmetrical counterparts of the visible region, and render the SMPL body using the reflected textures on the invisible region $\mathcal{T}_{\text{invisible}}$. The predicted invisible textures will then be used as a prior during optimization to serve as a stronger guide than SDS.

\paragraph{Overall Losses.} For the reference view $\beta_{\text{ref}}$, we directly minimize the per-pixel reconstruction loss between $\mathcal{I}$ and $\mathcal{G}_{\theta}(\beta_{\text{ref}})$, as well as the depth correlation loss \citep{tang2023make} between the rendered depth and the depth map (predicted by \citet{eftekhar2021omnidata}).
\begin{align}
    \mathcal{L}_{\text{ref view}} &= \mathcal{L}_{\text{rgb}} + \mathcal{L}_{\text{depth}} \\
   \mathcal{L}_{\text{rgb}} = || \mathcal{G}_{\theta}(\beta_{\text{ref}}) - \mathcal{I}|| &; \quad 
    \mathcal{L}_{\text{depth}} = - \frac{\text{Cov}(d(\beta_{\text{ref}}), d)}{\text{Var}(d(\beta_{\text{ref}})) \text{Var}(d)}
\end{align}
For a novel view $\beta_{\text{novel}}$, we minimize the SMPL-depth-conditioned SDS loss and 
\begin{align}
    \mathcal{L}_{\text{novel view}} &= \mathcal{L}_{d-SDS} + \mathcal{L}_{\text{invisible-RGB}} + \mathcal{L}_{\text{SMPL-depth}} + \mathcal{L}_{CLIP-D}
\end{align}
where
\begin{align}
    \mathcal{L}_{\text{invisible-RGB}} &= || \mathcal{R}(\mathcal{M}, \beta; \mathcal{T}_{\text{invisible}}) \odot m_{\text{inv}} - \mathcal{I} \odot m_{\text{inv}}||  \\
    \mathcal{L}_{\text{SMPL-depth}} &= - \frac{\text{Cov}(\mathcal{R}_d(\mathcal{M}, \beta), d)}{\text{Var}(\mathcal{R}_d(\mathcal{M}, \beta)) \text{Var}(d)}
\end{align}
Here $m_{\text{inv}}$ is the mask for the invisible region whose symmetrical counterpart is visible, and $\mathcal{R}(\mathcal{M}, \beta; \mathcal{T})$ is a mesh renderer that renders mesh $\mathcal{M}$ with texture $\mathcal{T}$ from view $\beta$. $\mathcal{L}_{CLIP-D}$ is a CLIP consistency loss \cite{tang2023make} that enforces semantic consistency between the rendered view and the reference image.
\begin{figure}[t]
    \centering
    \includegraphics[width=\textwidth]{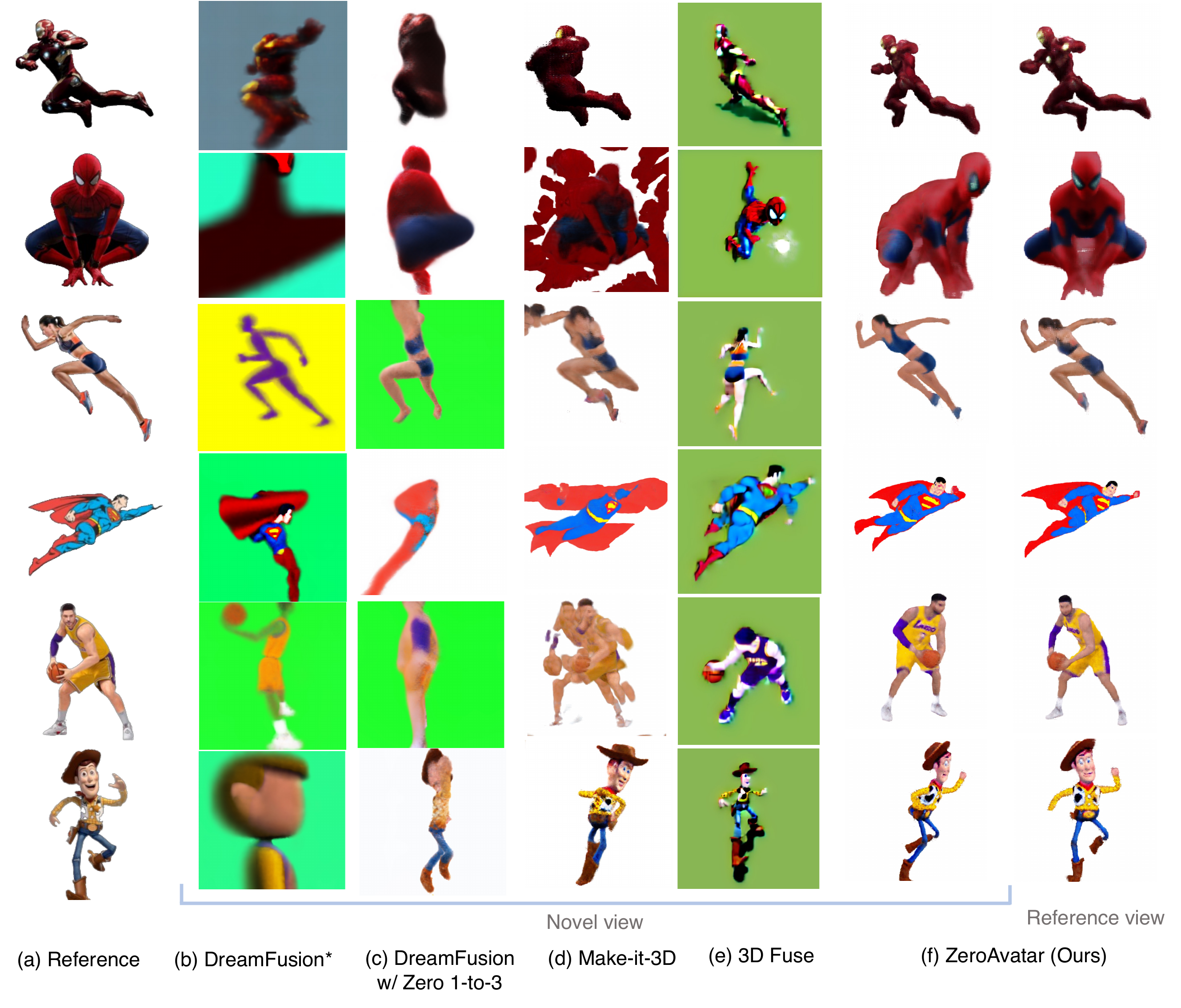}
    \caption{Comparison to state-of-the-art image-conditioned zero-shot optimization methods. \Ours{} demonstrates superiority over baselines Zero 1-to-3 \citep{liu2023zero} Make-it-3D \citep{tang2023make} where the underlying 3D geometry is not modeled explicitly. Although 3D Fuse \citep{seo2023let} approximates the 3D geometry using a point cloud, it still encounters the Janus problem ($3_{rd}$ row). Overall, \Ours{} yields results with higher fidelity and maintains greater consistency with the reference image.}
    \vspace{-4mm}
    \label{fig:qualitative_results}
\end{figure}

\section{Experiments}
\textbf{Implementation details.}
For efficiency, we adopt the multi-resolution hash encoding from Instant-NGP \citep{muller2022instant}, and follow several design choices from \cite{poole2022dreamfusion}, such as view sampling and shading augmentation. We use DensePose \citep{guler2018densepose} for UV coordinate regression, and PIXIE \cite{feng2021collaborative} for human mesh recovery. PIXIE outputs parameters of SMPL-X \citep{SMPL-X:2019} model, and we convert them into SMPL model parameters by finding the SMPL parameters that result in best mesh fit. NeRF rendering resolution during training is 100 by 100, and field of view is set to 20 degrees. We adopt progressive training following prior work \citep{tang2023make}, where we start with a narrow range of 90 degrees view symmetrical around the reference view and then gradually expand the range to cover 360 degrees during training. The narrow view is trained for 1,000 iterations, and the wide view is trained for 5,000 iterations. We turn off $\mathcal{L}_{\text{invisible-RGB}}$ and $\mathcal{L}_{\text{SMPL-depth}}$ after 2,000 iterations to avoid over-regularization of the geometry and appearance. We use Adam \citep{kingma2014adam} optimizer with learning rate $0.001$, and optimize for 6,000 iterations for a single image, which takes about 50 minutes on a single NVIDIA TITAN RTX GPU.

\textbf{Evaluation and metrics.}
Our method demonstrates greatest applicability for reconstructing out of distribution humans and virtual avatars that are difficult to curate ground truths for, and therefore unfeasible for learning-based methods. For this reason, we assess the performance of \Ours{} on a set of 27 images that include a wide variety of virtual avatars as well as challenging real-world humans. The persons from the evaluation set have a wide range of appearances, body poses and orientations. In Table \ref{table:quantitative}, we report Learned Perceptual Image Patch Similarity (LPIPS) \citep{zhang2018unreasonable}, contextual loss \cite{mechrez2018contextual}, and CLIP \cite{radford2021learning} similarity scores. LPIPS is computed on reference views. Contextual loss and CLIP similarity scores are computed on 360 degree views around the human, with each view positioned at 45-degree intervals from one another. CLIP similarity measures the semantic similarity between the novel view and the reference image (or text). Specifically, CLIP-I is the (normalized cosine) similarity score between the novel view and reference view computed using CLIP image embeddings, and CLIP-T is the similarity between novel view image embeddings and text (i.e. image captions) embeddings. We calculate all metrics for each test scene, and report the average over all test scenes.

\begin{table}[ht]
\vspace{-4mm}
\caption{Quantitative results. Best numbers are in \textbf{bold}. ($^{*}$: Stable–Dream Fusion$^{*}$ is a variant of DreamFusion that that additionally minimizes reconstructions loss for reference view every few iterations.)}
\centering
\small
\resizebox{\textwidth}{!}{
\begin{tabular}[t]{@{} l *4c @{}}
\toprule
\multicolumn{1}{c}{Model} & LPIPS ($\downarrow$) &  Contextual ($\downarrow$) & CLIP-I ($\uparrow$) & CLIP-T ($\uparrow$) \\ 
\midrule
Stable–DreamFusion$^{*}$ \cite{poole2022dreamfusion} & 0.596 & 3.55 & 73.55  & 26.64 \\
Stable-DreamFusion (w. Zero 1-to-3) \cite{liu2023zero} & 0.391 & 3.35 & 76.19 & 27.05 \\
3D Fuse \cite{seo2023let} & 0.704 & 3.20 & 81.07 & 30.50\\
Make-it-3D - Coarse \citep{tang2023make} & \textbf{0.342} & 3.35 & 85.86 & 32.00 \\
\Ours{}  (Ours) & 0.359 & \textbf{3.15} & \textbf{87.67} & \textbf{32.59} \\
\bottomrule
\end{tabular}
}
\vspace{-2mm}
\label{table:quantitative}
\end{table}



\textbf{Results.
}
We compare \Ours{} to state-of-the-art zero-shot methods (Figure \ref{fig:qualitative_results}), namely DreamFusion \citep{poole2022dreamfusion}, Zero 1-to-3 \citep{liu2023zero}, 3D Fuse \citep{seo2023let}, and Make-it-3D \citep{tang2023make}. For DreamFusion\cite{poole2022dreamfusion} we use the open-sourced implementation Stable-DreamFusion \citep{stable-dreamfusion}. DreamFusion$^{*}$ (column b) is an image-conditioned variant of DreamFusion that additionally minimizes reconstructions loss for reference view every few iterations. In column c, we use pre-trained Zero 1-to-3 model from \citet{liu2023zero} as supervision to optimize the neural radiance field. Zero 1-to-3 is trained on a large-scale open-source dataset containing 800K+ 3D models \citep{deitke2022objaverse}, and learns good prior of how an object should look like from novel view points. However, for humans in general, the novel views predicted by Zero 1-to-3 are not sufficient as supervision signal. 
\begin{wrapfigure}[21]{r}{0.55\textwidth}
    \includegraphics[width=0.55\textwidth]{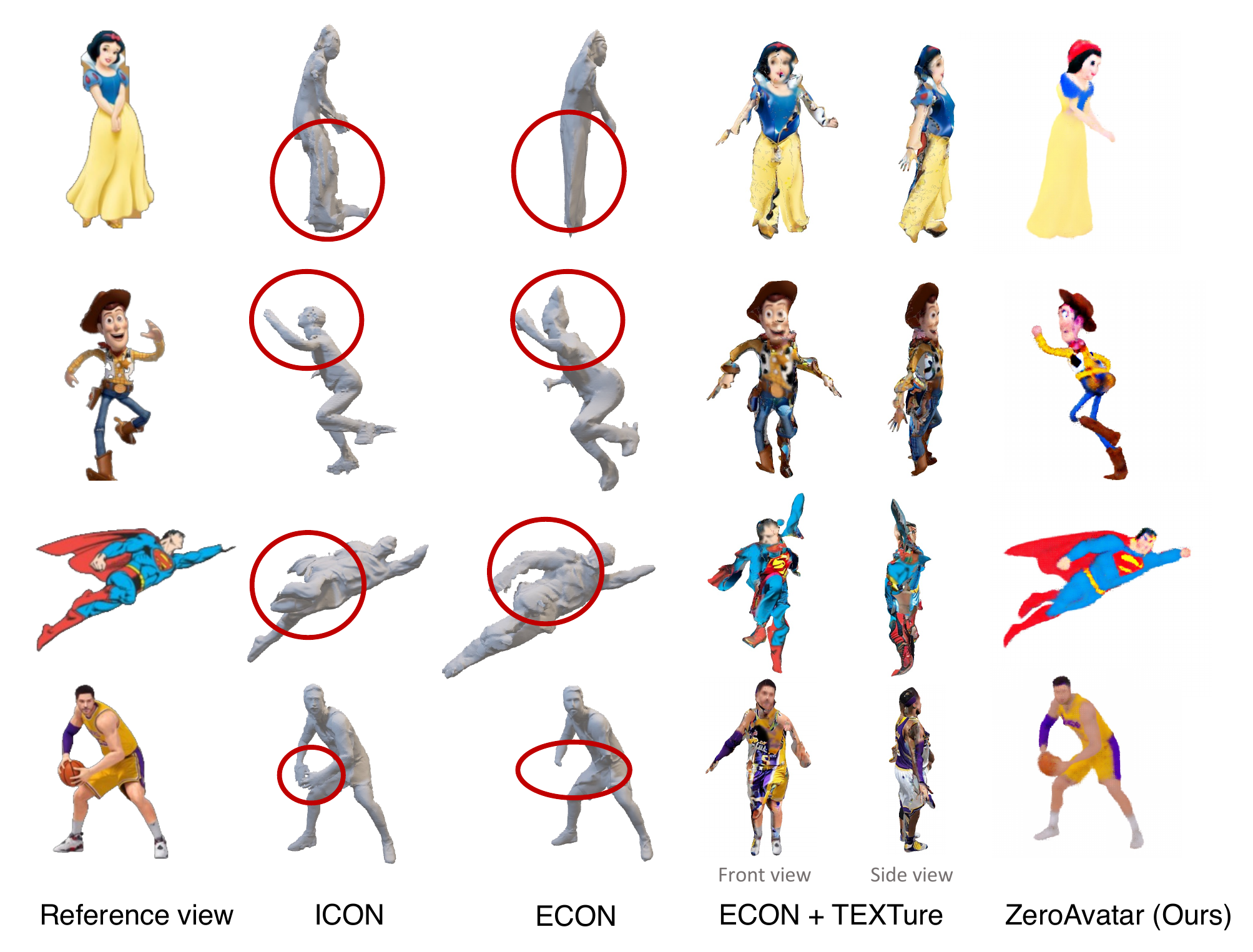}
    \caption{Comparison to learning-based methods: ICON \citep{xiu2022icon} and ECON \citep{xiu2022econ}. Learning-based methods have limited generalization capability due to the limited variety of the training data. In comparison, \Ours{} as a zero-shot method is able to recover rare details such as cape (3rd row) and basketball (4th row).}
    \label{fig:learning}
\end{wrapfigure}
Make-it-3D \citep{tang2023make} (column d) is a recent state-of-the-art image-conditioned zero-shot method that has two stages of training. We compare to the results after first stage since code for second stage is not released as of now. However, we note that the geometry of the object is mainly learned in their first stage, and second stage is for texture refining. Thus, comparing to results post Stage I should give us a good sense of how our learned geometry compares to theirs. We observe that Make-it-3D tends to learn reasonable geometry for humans that have simple geometry (e.g. a near cylindrical volume as in the bottom row). For humans with more difficult poses, however, the geometry quickly diverges for novel views and score distillation loss with text conditioning fails to correct the geometry. The degenerate geometry from novel views results in higher contextual loss and lower CLIP scores.

3D Fuse \citep{seo2023let} (column e) incorporates 3D awareness in the form of a coarse point cloud, which enhances the robustness and 3D consistency of learned neural radiance field. We use the image-to-3D version of 3D Fuse for comparison. Since explicitly modelling the underlying shape, 3D Fuse yields much better geometry as compared to DreamFusion and Make-it-3D. However, as shown the results still frequently encounter the Janus problem (where the learned 3D model has multiple faces) when it comes to humans with complex poses (e.g. 3rd row). In addition, since 3D Fuse does not operate on image-level reconstruction loss, the output model does not align well with the reference image. In comparison, \Ours{} consistently produces more realistic avatars that are more consistent with the input image.

Quantitatively (Table \ref{table:quantitative}), \Ours{} consistently outperforms DreamFusion, DreamFusion with Zero 1-to-3, and 3D Fuse \cite{poole2022dreamfusion,liu2023zero,seo2023let} across all metrics, which demonstrate its effectiveness and superiority in comparison to these baselines. \Ours{} achieves similar metrics as Make-it-3D \cite{tang2023make}. However, we note that for some test images, while quantitative metrics are close or even lower for \Ours{}, the qualitative results for novel views show significant improvement over Make-it-3D \cite{tang2023make}, where learned geometry tends to over-fit to reference view and drastically distorts when viewed from novel angles (Figure \ref{fig:qualitative_results}). To further showcase that our method is better at preserving human structure, we use the detection score from an off-the-shelf human detector \cite{girshick2015fast} on novel views as a proxy for evaluating the structural integrity of human subjects. On average, novel views from DreamFusion (image-conditioned or with Zero 1-to-3) both have lower than 50$\%$ detection score. Make-it-3D and 3D Fuse attain $67\%$ and 74$\%$ respectively, and \Ours{} achieves 83$\%$. This indicates that \Ours{} is better at preserving the structural integrity of human subjects as compared to alternative approaches.

Additional qualitative results and animated results can be found in \textit{Supplemental Materials}.

\begin{wrapfigure}[23]{r}{0.33\textwidth}
  \centering
  \begin{subfigure}[b]{0.33\textwidth}
    \centering
    \includegraphics[width=\textwidth]{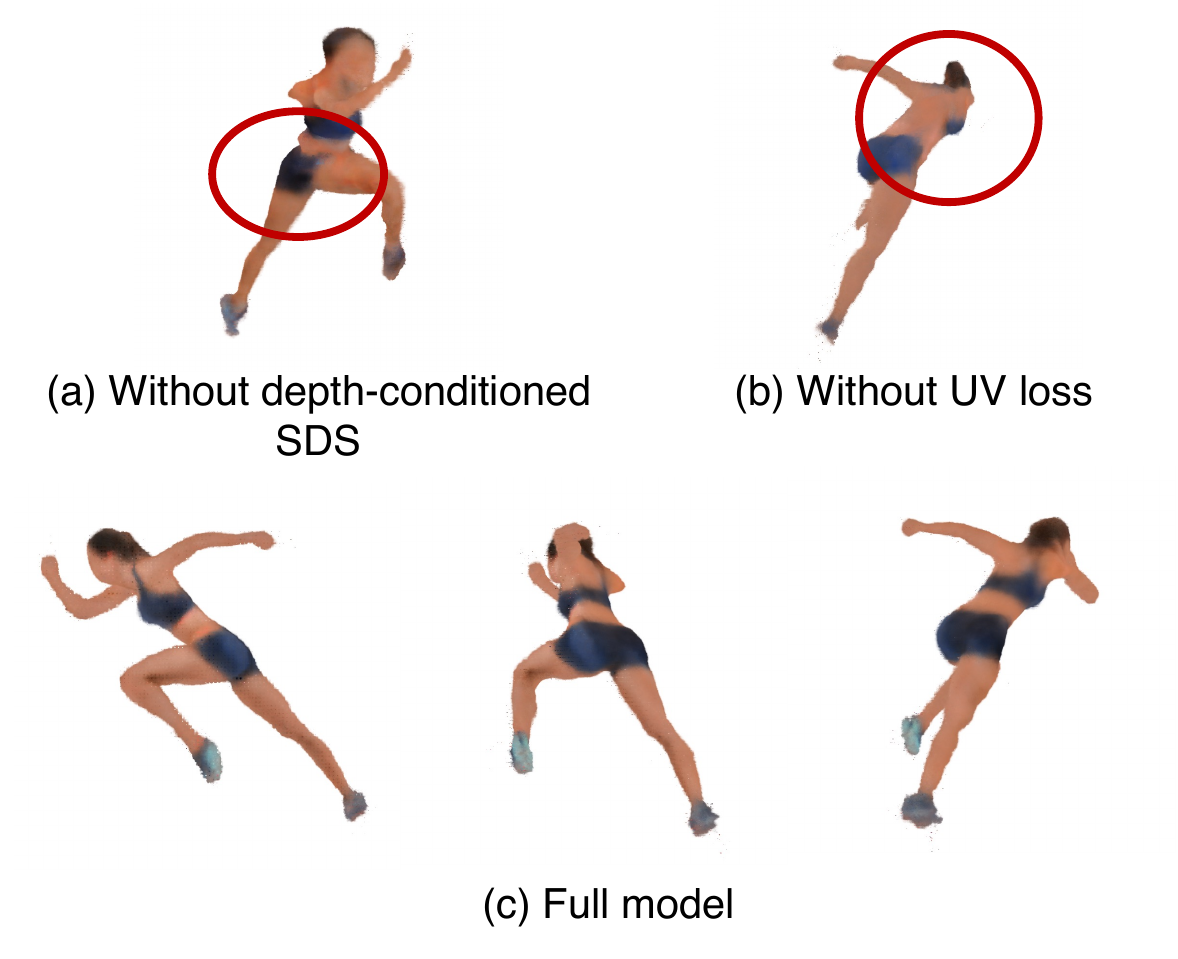}
    \caption{Ablated models.}
    \label{fig:ablations}
  \end{subfigure}
  \vfill
  \begin{subfigure}[b]{0.33\textwidth}
    \centering
    \includegraphics[width=\textwidth,height=3cm]{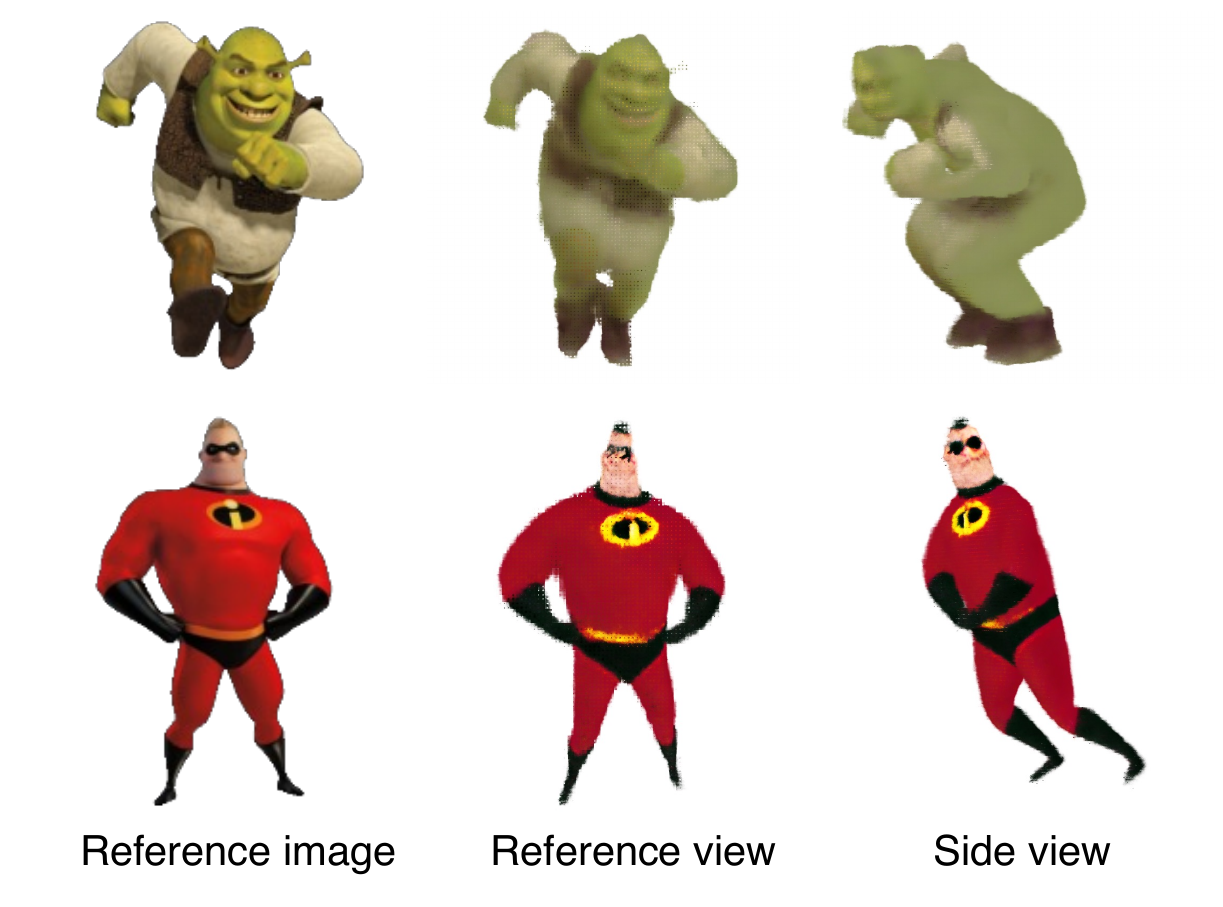}
    \caption{Failure cases.}
    \label{fig:limitations}
  \end{subfigure}
  \caption{Ablations and limitations.}
  \label{fig:twosubfigs2}
\end{wrapfigure}

\textbf{Comparison to learning-based methods.} There is a line of research that directly estimates volumetric representations of humans from single-view images. We compare to ICON \citep{xiu2022icon} and ECON \citep{xiu2022econ} (Figure \ref{fig:learning}). Additionally, we include the textures optimized by TEXTure \citep{richardson2023texture} using the mesh predicted by ECON. Since ICON and ECON are trained end-to-end on 3D scans such as Renderpeople \citep{Renderpeople_2022}. Although they achieve impressive results on comparable images (i.e. real humans), they demonstrate limited generalization capability to virtual avatars such as cartoon characters and out-of-distribution scenarios (e.g. basketball player holding a ball). As shown in Figure \ref{fig:learning}, \Ours{} is better at generalization. In the case of real humans (4th row), \Ours{} shows superior capability at capturing intricate details and displaying better geometry overall.

\textbf{Model ablations.}
We demonstrate the effect of each component of our method in Figure \ref{fig:ablations}. First, we only initialize the density field with the occupancy of SMPL body mesh, but do not use the depth values of SMPL for SDS. 
As shown in Figure \ref{fig:ablations} (a), the learned body often exhibits unrealistic anatomy. Second, we take out the UV prior loss and observe that the appearance of the non-visible side of the person is less realistic than our full model.

\begin{figure}[h!]
  \centering
    \includegraphics[width=\textwidth]{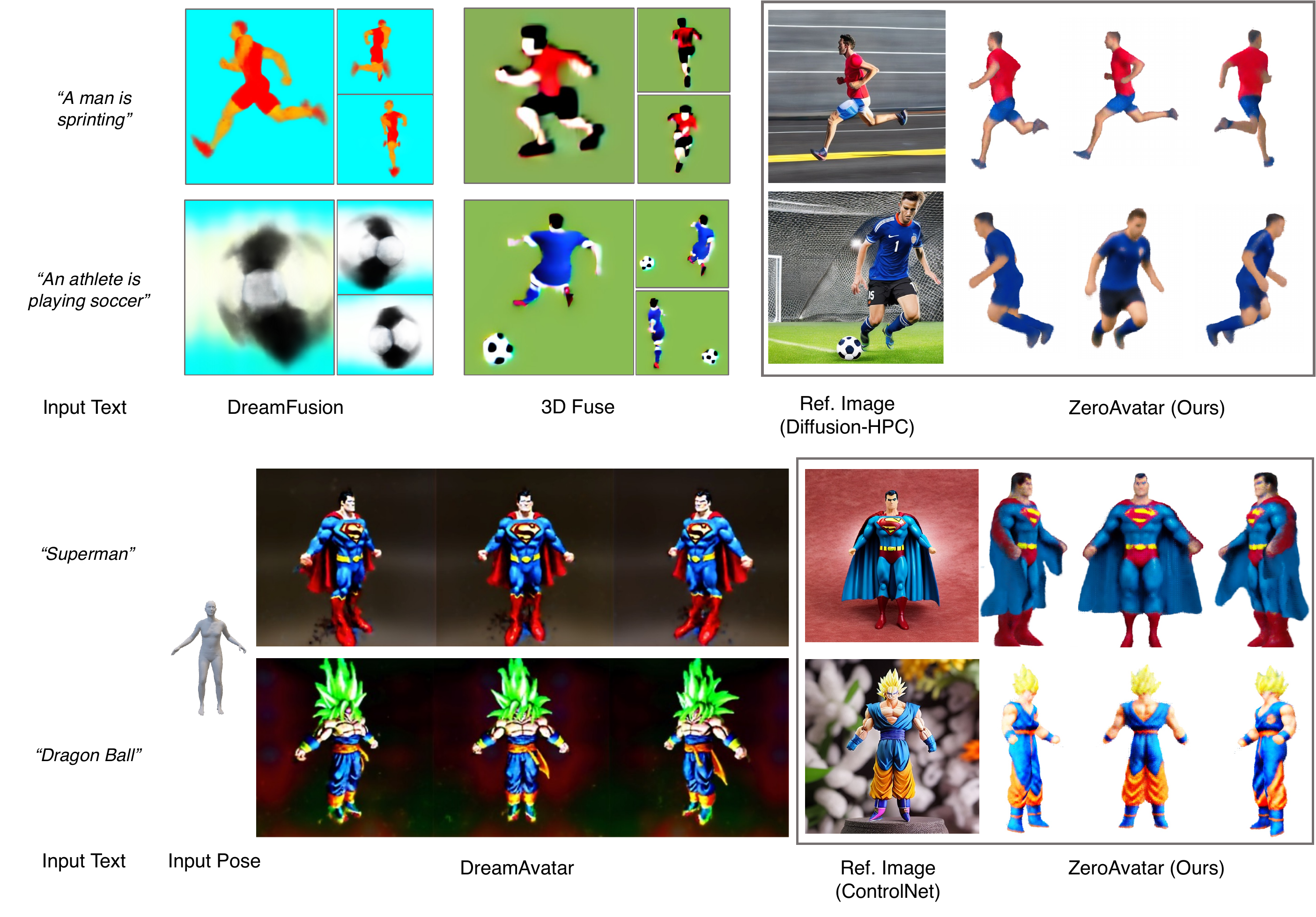}
    \caption{Application of \Ours{} in text-to-3D avatar generation using text-to-image model as a stepping stone. Top: Text-to-3D generation. Bottom: Pose-conditioned text-to-3D generation. Compare to DreamFusion \cite{poole2022dreamfusion}, 3D Fuse \cite{seo2023let} and DreamAvatar \cite{cao2023dreamavatar} that directly optimize a NeRF from text, our method can efficiently reconstruct a high-fidelity 3D model from text, using synthetic image as an intermediate representation. }
    \label{fig:application}
\end{figure}

\textbf{Application: zero-shot text-to-3D avatar generation.}
\Ours{} can be used with text-to-image models to achieve zero-shot text-to-3D capability with optional pose control. We showcase such applications using Diffusion HPC \citep{weng2023diffusion} and ControlNet \citep{zhang2023adding}. 
In top half of Figure \ref{fig:application}, given a text prompt, we use Diffusion HPC \citep{weng2023diffusion}, to generate a synthetic image as the reference image, and then apply \Ours{} on the reference image. As shown, as compared to baselines DreamFusion and 3D Fuse that directly optimize the 3D representation given a text prompt, our result attains much higher fidelity due to the usage of an intermediate photo-realistic synthetic image.

For pose-conditioned text-to-3D generation (bottom half of Figure \ref{fig:application}), we use ControlNet \citep{zhang2023adding} to generate a synthetic image given the input pose in the form of 2D keypoints. We compare to DreamAvatar \cite{cao2023dreamavatar}, a SDS-based method that adopts a dual-space (i.e. canonical and observed pose space) optimization scheme to regularize the shape of the person, which incurs additional computational costs and takes about 2 hours on a GPU to optimize a single model. In comparison, \Ours{} with text-to-image generation as stepping stone alleviates the need for further regularization, and achieves comparable or better fidelity with much less time (about 50 minutes).

\textbf{Limitations and future directions.} \Ours{} assumes that the geometry of the human avatar in the reference image can be approximated by a parameterized body model (i.e. SMPL). The SMPL model is designed to capture the average shape and pose variations of human bodies. Thus, in scenarios when the proportion of the human deviates too much from the shape space represented by SMPL, the resulting estimation may lead to disproportionate representations (Figure \ref{fig:limitations}). A promising avenue for future research could be enhancing the generalizability of the human body prior. In addition, despite \Ours{}'s superior capability of preserving human geometry, the 3D mesh extracted from the learned density fields are still relatively coarse. Nonetheless, our work could be synergistically combined with other approaches  \citep{chen2023fantasia3d} that refine geometry and texture resolution, thereby further improving the visual appearance and geometric fidelity.

\vspace{-4mm}
\section{Conclusion}
In this work, we proposed \Ours{}, a zero-shot method for creating high-fidelity 3D avatars from a single image, using a pre-trained text-to-image diffusion model as a prior. \Ours{} significantly enhances the robustness
and ensure 3D consistency of optimization-based image-to-3D avatar generation. On images of posed humans, \Ours{} surpasses existing zero-shot methods in terms of the optimized geometry and appearance. Further, \Ours{} can be seamlessly combined with a pre-trained text-to-2D method, enabling the generation of 3D avatars with text or pose control, allowing for a wide range of usage scenarios and creative possibilities.

\textbf{General impact. } \Ours{} offers a efficient way for content creators to generate 3D human avatar models with image, text or pose control. Due to the usage of a pre-trained text-to-image model, our approach inherits any biases and limitations associated with it. There is also potential risk of generating 3D models of individuals without their consent. We recommend that users adhere to proper usage practices.
\section*{Appendix}

\subsection*{Effect of Pose Complexity}
We observe that qualitatively, \Ours{} exhibits the most significant improvement compared to the baselines, particularly when the pose deviates significantly from the canonical pose (hence being more ``complex"). In order to further understand how poses impact the optimization results, we perform evaluation on poses with varying complexity. Specifically, we use VPoser \cite{SMPL-X:2019}, a pre-trained pose prior, to quantify the complexity of the poses, which we define as the amount of deviation from the canonical pose in the VPoser embedding space. Formally, complexity score is computed as
\begin{align}
    \text{score} &= \text{mean}(\mathcal{E}_{\text{VPoser}} (\Omega)^2)
\end{align}
where $\Omega$ is the pose of the person, and $\mathcal{E}_{\text{VPoser}}$ is the encoder of VPoser. A higher score indicates a more complex pose. We then bin the test cases into easy, medium and hard categories based on the pose complexity score. Easy poses are those that fall within the 50th percentile, medium poses are within the 75th percentile, and hard poses encompass the remaining cases. In Table \ref{table:supp} we report the quantitative results on binned categories. We observe that \Ours{} shows consistent improvement for different types of poses over baselines in terms of contextual loss and CLIP similarity scores. Although we sometimes attain worse LPIPS, we observe that it is because baselines tend to over-fit to the reference view at the price of distorting the overall geometry (e.g. Figure \ref{fig:supp_3}). 

\begin{table}[ht]
\caption{Quantitative results on test cases with easy / medium / hard poses. }
\centering
\small
\resizebox{\textwidth}{!}{
\begin{tabular}[t]{@{} l *4c @{}}
\toprule
\multicolumn{1}{c}{Model} & LPIPS ($\downarrow$) &  Contextual ($\downarrow$) & CLIP-I ($\uparrow$) & CLIP-T ($\uparrow$) \\ 
\midrule
Stable-DreamFusion$^{*}$ \cite{poole2022dreamfusion} & 0.60 / 0.51 / 0.67 & 3.8 / 3.4 / 3.6 & 77.1 / 80.3 / 76.4  & 28.0 / 29.0 / 29.0 \\
Stable-DreamFusion (w. Zero 1-to-3) \cite{liu2023zero} & 0.42 / \textbf{0.33} / 0.44 & 3.5 / 3.3 / 3.3 & 76.9 / 77.1 / 74.6 & 27.7 / 27.0 / 26.6 \\
3D Fuse \cite{seo2023let} & 0.68 / 0.67 / 0.75 & 3.3 / 3.1 / 3.2 & 80.7 / 84.1 / 83.4 & 30.5 / 31.6 / 32.1  \\
Make-it-3D - Coarse \citep{tang2023make} & \textbf{0.37} / 0.34 / 0.39 & 3.4 / 3.3 / 3.3 & 83.7 / 89.7 / 85.2 & 32.1 / 33.5 / 32.0 \\
\Ours{}  (Ours) & 0.38 / 0.34 / \textbf{0.37} & \textbf{3.3} / \textbf{3.0} / \textbf{3.2} & \textbf{85.0} / \textbf{89.7} / \textbf{87.0} & \textbf{32.2} / \textbf{33.8} / \textbf{32.0} \\
\bottomrule
\end{tabular}
}
\label{table:supp}
\end{table}

\subsection*{Additional Qualitative Reults}
\begin{figure}
    \centering
    \includegraphics[width=0.9\textwidth]{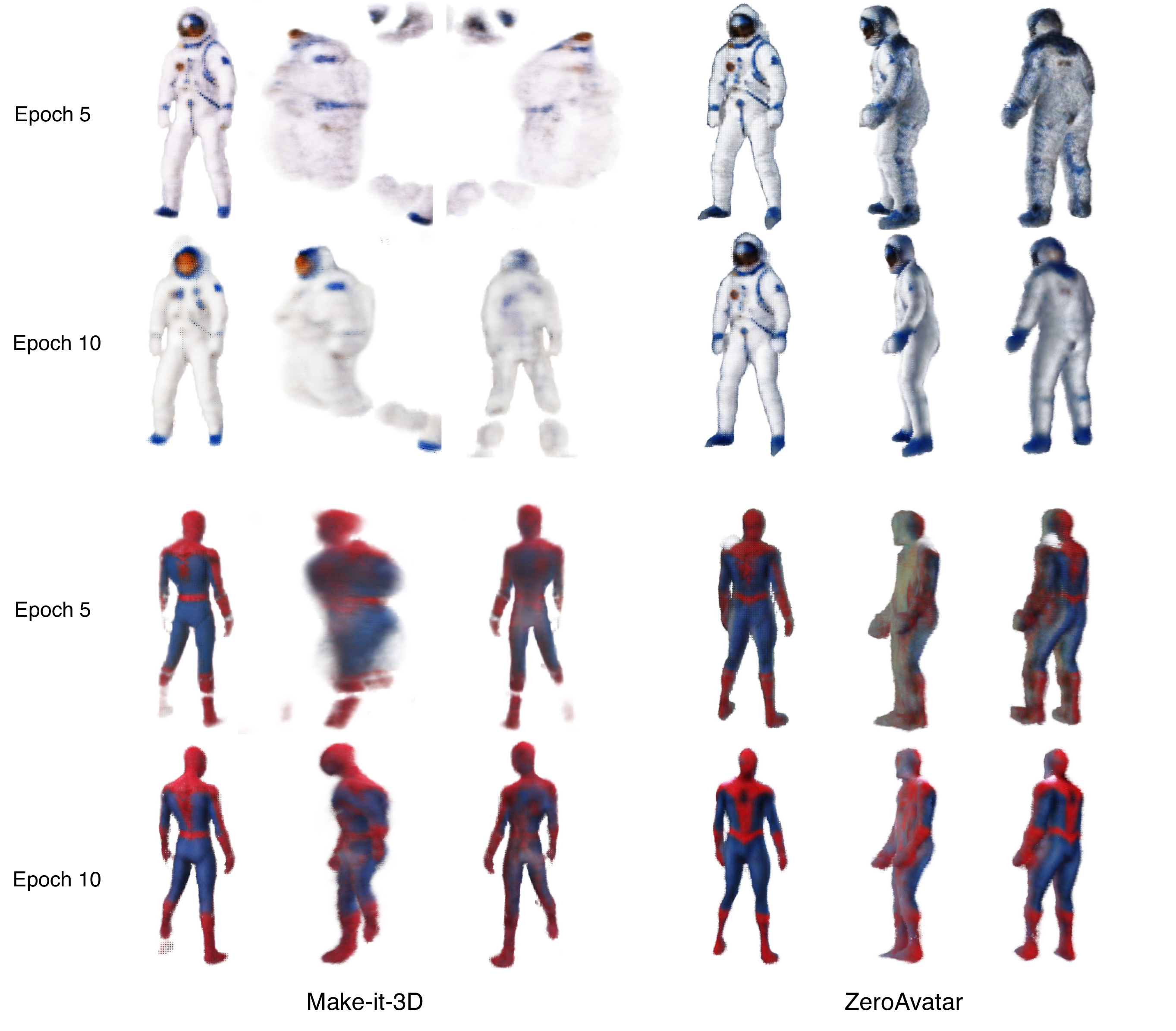}
    \caption{Renderings of learned NeRFs using Make-it-3D \cite{tang2023make} and \Ours{} at end of epoch  5 and 10.}
    \label{fig:supp_1}
\end{figure}

To visually showcase the progress of optimization, we showcase renderings from 3 viewpoints at end of epoch 5 and 10 (Figure \ref{fig:supp_1}). As shown, \Ours{} has much faster convergence speed. Since it uses a coarse geometry as initialization, the optimized NeRF shows roughly correct geometry and colors as early as epoch 5, whereas baseline Make-it-3D takes at least 10 epochs to attain a bounded shape.

\begin{figure}
    \centering
    \includegraphics[width=0.9\textwidth]{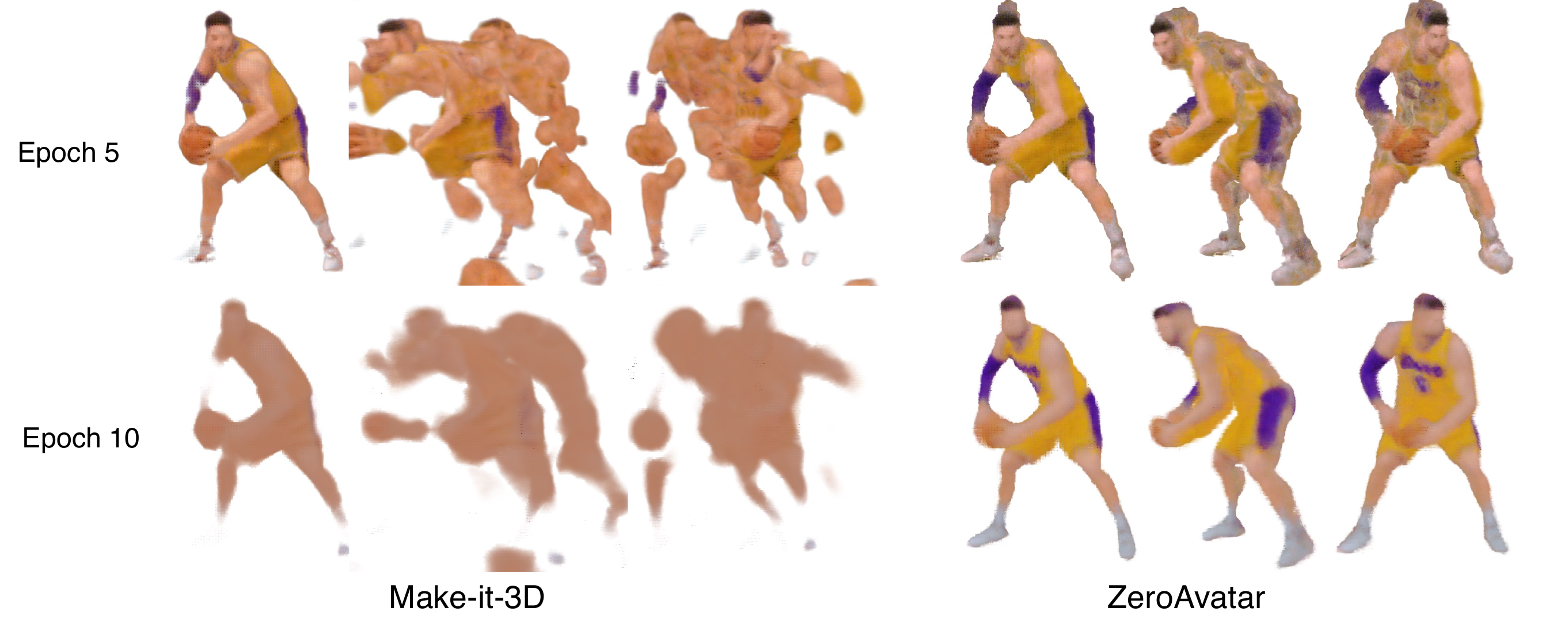}
    \caption{Make-it-3D \cite{tang2023make}'s optimization sometimes results in divergence, whereas \Ours{} demonstrates more robust optimization.}
    \label{fig:supp_2}
\end{figure}

In addition, we observe that Make-it-3D occasionally diverges (Figure \ref{fig:supp_2}) after which point optimizing for reference view reconstruction loss shows little effect in rectifying the geometry of the person. In comparison, \Ours{} consistently produces more robust results.

\begin{figure}
    \centering
    \includegraphics[width=0.9\textwidth]{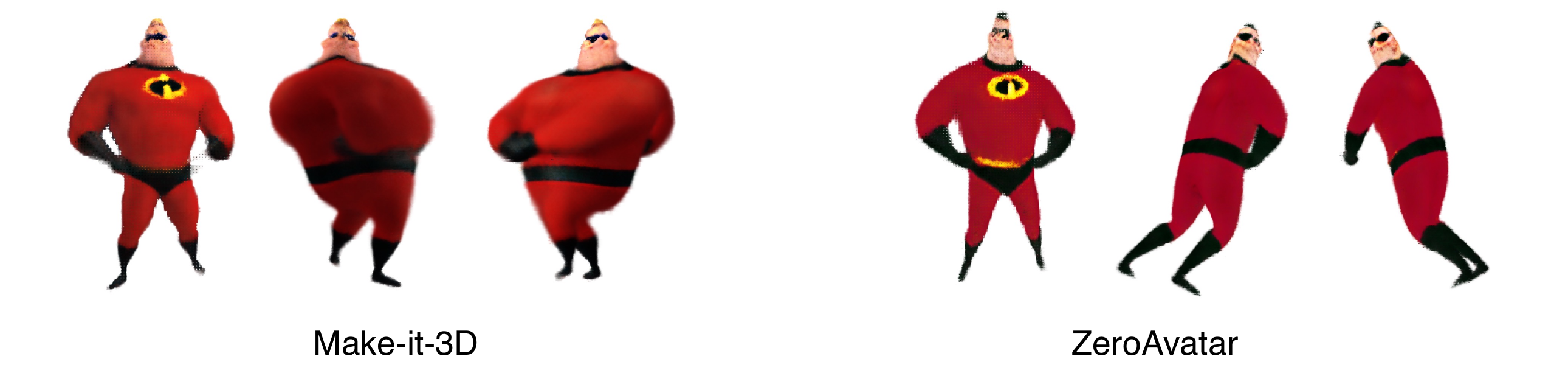}
    \caption{Failure scenario where SMPL cannot faithfully represent the proportion of the avatar. Nonetheless, \Ours{} still produces novel view renderings that are more consistent with reference view as compare to baseline (Make-it-3D).}
    \label{fig:supp_3}
\end{figure}

Lastly, we include example for a failure case (Figure \ref{fig:supp_3}) where the person's shape cannot be accurately represented by SMPL body model \cite{loper2015smpl}. As shown, although \Ours{} has slight misalignment with the input image, the shape of the person from novel views are more consistent with the reference view, whereas novel view renderings from Make-it-3D often get distorted.

\medskip

\newpage
{\small
\bibliographystyle{abbrvnat}
\bibliography{egbib}
}
\end{document}